\documentclass[10pt,twocolumn,letterpaper]{article}

\usepackage{cvpr}
\usepackage{times}
\usepackage{epsfig}
\usepackage{graphicx}
\usepackage{amsmath}
\usepackage{amssymb}

\usepackage{multirow}
\usepackage{makecell}

\usepackage[breaklinks=true,bookmarks=false]{hyperref}

\cvprfinalcopy 

\def\cvprPaperID{****} 

\begin{document}

\title{Learning to Synthesize Fashion Textures}

\author{Wu Shi\\
The Chinese University of Hong Kong\\
{\tt\small sw015@ie.cuhk.edu.hk}
\and
Tak-Wai Hui\\
The Chinese University of Hong Kong\\
{\tt\small twhui@ie.cuhk.edu.hk}
\and
Ziwei Liu\\
The Chinese University of Hong Kong\\
{\tt\small zwliu.hust@gmail.com}
\and
Dahua Lin\\
The Chinese University of Hong Kong\\
{\tt\small dhlin@ie.cuhk.edu.hk }
\and
Chen Change Loy\\
Nanyang Technological University\\
{\tt\small ccloy@ntu.edu.sg}
}

\maketitle

\begin{abstract}
Existing unconditional generative models mainly focus on modeling general objects, such as faces and indoor scenes.
Fashion textures, another important type of visual elements around us, have not been extensively studied.
In this work, we propose an effective generative model for fashion textures and also comprehensively investigate the key components involved: internal representation, latent space sampling and the generator architecture.
We use Gram matrix as a suitable \textit{internal representation} for modeling realistic fashion textures, and further design two dedicated modules for modulating Gram matrix into a low-dimension vector.
Since fashion textures are scale-dependent, we propose a \textit{recursive auto-encoder} to capture the dependency between multiple granularity levels of texture feature.
Another important observation is that fashion textures are multi-modal. We fit and sample from a \textit{Gaussian mixture model} in the latent space to improve the diversity of the generated textures.
Extensive experiments demonstrate that our approach is capable of synthesizing more realistic and diverse fashion textures over other state-of-the-art methods.
\end{abstract}

\section{Introduction}
\label{sec:introduction}

\begin{figure}
  \centering
  \includegraphics[width=0.8\linewidth]{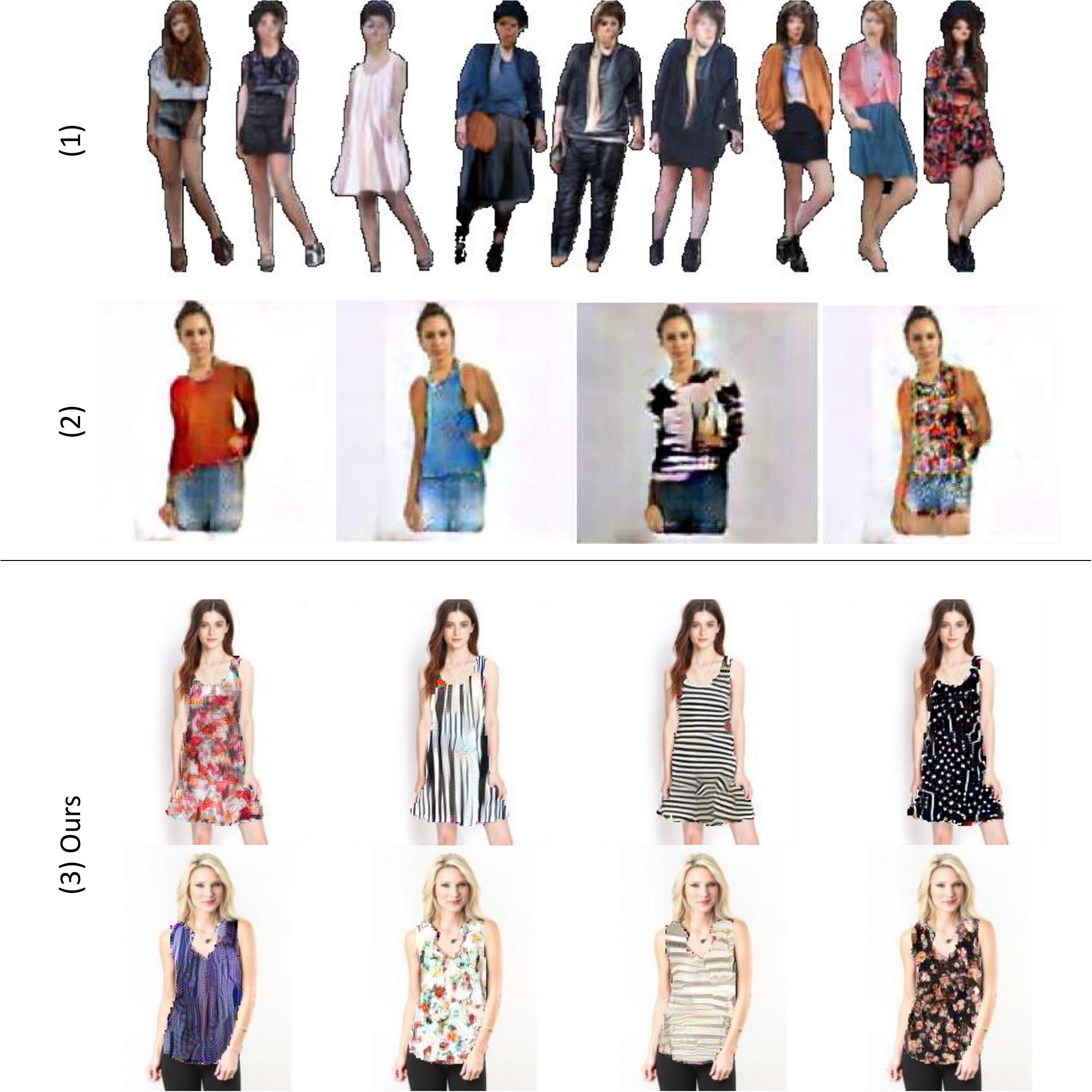}\\
  \caption{The results of generative methods for fashion clothing: (a) Lassner \etal~\cite{lassner2017generative}, (b) Zhu \etal~\cite{zhu2017your}, and (c) ours. Both (a) and (b) learn to generate an image directly, while our method first learns to generate a set of Gram matrices and then generate a texture image using that feature. The results of (a) and (b) are copied from their original papers. The resolution of our method is $256\times 256$. More details can be seen by enlarging the images on a color display.}\label{fig:overview}
\end{figure}

%

Fashion texture can be frequently seen in clothing items and constitutes an important element in our visual world. It also plays an important role in human-centered generation tasks. Existing generative models mainly focus on modelling general objects like faces~\cite{karras2017progressive} and body shapes~\cite{lassner2017generative}. Generation for fashion textures, however, has not been studied extensively. In the context of fashion synthesis, existing methods often render clothing images conditioned on various clues including body-part segmentatoins~\cite{lassner2017generative}, style descriptions~\cite{zhu2017your}, texture patches~\cite{xian2018texturegan}, clothing items~\cite{han2018viton}, or people in clothings~\cite{raj2018swapnet}. In contrast, we aim to design an unconditional generative model for fashion textures, which can generate realistic and diversified texture images. Learning a generative model for textures has benefits to reducing the need of using massively annotated data to learn for texture models and improving robustness to different tasks and contexts~\cite{kingma2018glow}. The proposed method is an independent and flexible module that can be closely integrated with other synthesis methods.

Fashion textures have specific statistic properties that can be characterized as realizations of a stationary, ergodic and Markovian process~\cite{georgiadis2013texture}. Such characteristics should be deliberated in order to generate realistic and high-quality textures.
Existing fashion generative methods~\cite{lassner2017generative,zhu2017your} usually follow the common settings, \eg training a convolutional network with pixel-wise reconstruction loss and/or adversarial loss in image space. As shown in Figure~\ref{fig:overview} (a-b), the generated clothes contain well-defined shapes and smoothed regions but detailed textures are often missing. Generation of realistic fashion textures, especially from scratch, remains a challenging task in the literature.
The challenges include:
1) Textures have self-correlated structures and they require high-dimensional and high-order statistics to describe in both the traditional texture analysis literature~\cite{julesz1962visual} and recent deep learning based texture synthesis literature~\cite{gatys2015texture}.
2) Deep learning based texture generation methods need to constrain multiple levels of features in the texture model. Such features are usually extracted from different layers of a neural network. When generating using only the lowest level, the textures have little structure which are similar to noise with matched color. While increasing the number of used levels will increase the degree of structure for generated textures~\cite{gatys2015texture}. Moreover, the texture features from different levels should be consistent with each other, otherwise the generated images may contain inconsistent and unnatural patterns (Figure~\ref{fig:ablation_study} (b)).
3) Textures are multi-modal and not homogeneously spanned in the feature space. Inappropriate training and inferencing methods can result in unrealistic samples (which are dissimilar to the training data) and small diversity of samples (known as mode collapse in GAN) (Figure~\ref{fig:ablation_study} (c)).

To address the above challenges for more realistic and diverse fashion texture synthesis, we identify the following key components for a successful generative model:
\begin{enumerate}
  \item We choose the Gram matrix (the second order moments) of the activation of non-linear convolutional network filters, as the texture feature and train our model to generate that feature explicitly. Employing Gram matrix is the central idea in many state-of-the-art deep learning methods for texture synthesis~\cite{gatys2015texture,ulyanov2016texture} and style transfer~\cite{johnson2016perceptual,li2017universal}. It is shown to be a powerful representation that can specify the spatial summary statistics of texture features. It is also a stationary representation and can be computed in almost arbitrary texture regions, while other methods operating in image space need to deal with the variance in appearance of textures. We further propose a dedicated transform layer for structure-preserving dimension reduction and efficient computation.
  \item We propose a recursive auto-encoder network structure, which consists of a stack of recursive units built on multi-level Gram matrices. This structure encodes and generates the input features level-by-level and can better model the correlation between different levels of features.
  \item In the inference stage, we sample from a Gaussian Mixture Model fitted in the latent space. Using this latent distribution can alleviate the multi-modality and non-homogeneous variance problems in the latent dimensions.
\end{enumerate}

Finally, we conduct a comprehensive evaluation of our approach on a significantly larger and more diversified fashion texture dataset comparing to the datasets used by previous works. The experimental results show that our approach outperforms other baseline methods by a large margin with respect to visual quality, distance metric (FID) and user preference.

\section{Related Work}
\label{sec:related_work}

\noindent\textbf{Generative Models for Fashion Items.} Current generative methods for fashion clothing are closely related to generating people in clothing. \cite{lassner2017generative} is the first one to generate clothing images from human body segmentation. \cite{zhu2017your} further uses language descriptions to generate stylized clothes. Although their methods can generate realistic shape of clothes, the texture of clothes are not good enough. As can be seen in Figure~\ref{fig:overview}, the generated images contain mostly smooth regions, while little or even no textures can be observed. \cite{date2017fashioning} applies neural style transfer algorithm to synthesize new custom clothes. Their method requires a set of reference images from user during sampling. The generated results are simply a mixed version of the reference images. Our method aims to generate realistic and diversified textures from scratch.

\noindent\textbf{Texture Generation.} The methods for texture generation can be roughly divided into two groups: 1) exemplar-based texture generation, and 2) non exemplar-based texture generation. The first group of texture generation methods requires an example texture image as the reference patch. Their goal is to generate texture images that share the similar texture feature (i.e. Gram matrices).~\cite{gatys2015texture,berger2016incorporating} formulate the synthesis process as a optimization problem and does not requires training.~\cite{ulyanov2016texture,johnson2016perceptual} employ the Gram matrices as a perceptual loss and their methods can achieve real-time sampling. The main drawback of this kind of method is that the generated results are not diverse and the generative model can not generate novel textures after being trained on several textures. Although the generated images may vary in the pixel-wise appearance, perceptually they looks similar to the reference texture patch.
Another group of texture generation methods do not require an example texture. Generally speaking, they work in a adversarial way to generate a group of texture images. \cite{jetchev2016texture} proposes to use a spatial GAN by extending the
input noise distribution space to suit to the texture generation task. \cite{bergmann2017learning} proposes to add a periodic dimension in the noise space to generate well structured textures. The limitation is that their assumptions are too strong and cannot model a wide range of textures.

\noindent\textbf{Generative Models} Recently VAE~\cite{kingma2013auto} and GAN~\cite{goodfellow2014generative} has drawn increasing interests in the context of image synthesis. Various techniques~\cite{salimans2016improved,gulrajani2017improved,tolstikhin2017wasserstein,dai2018diagnosing,miyato2018spectral,tran2018dist} are proposed to improve the quality, realism, diversity of generated results and the stablity of the training process. DistGAN~\cite{tran2018dist} is an improved GAN method that alleviates the gradient vanishing and mode collapse problems. This method can produce more diversified results than vanilla GAN. Like vanilla GAN, this method operate in the image space which is different from our setting. The recently proposed WAE~\cite{tolstikhin2017wasserstein} combines both advantages of VAE and GAN in a single framework. Through experiments we find that the WAE framework is quite suitable for our case, as it is stabler than GAN in the training process and produces better results than VAE does. Thus, we will employ this framework as our generative method throughout our experiments.

\section{Fashion Texture Generation}
\label{sec:method}

\begin{figure*}
  \centering
  \includegraphics[width=0.98\linewidth]{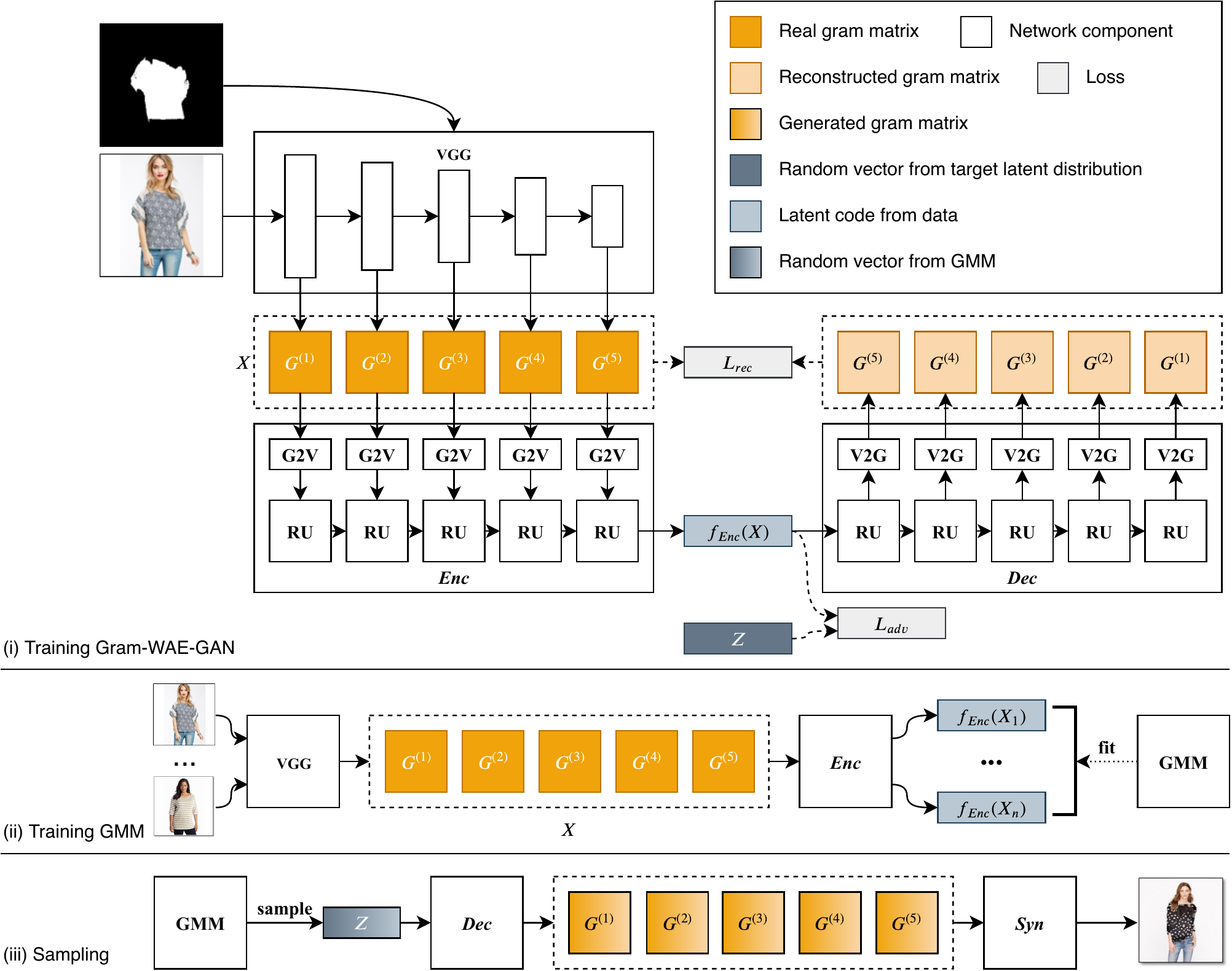}\\
  \caption{Training and sampling pipelines. In the training step (i), an auto-encoder, based on the WAE-GAN framework, is learnt to minimize the reconstruction loss of Gram matrices and the adversarial loss between the latent code $f_{Enc}(X)$ and the reference noise vector $Z$ sampled from $N(0,I)$. Two modules (\textbf{G2V} and \textbf{V2G} in Section~\ref{subsec:gram_matrix_transformation}) are designed for projecting Gram matrix into and from a low-dimension vector. A stack of recursive units (\textbf{RU} in Section~\ref{subsec:recursive_autoencoder}) is used in both the encoder and decoder networks to capture the dependency between multiple granularity levels of texture feature. In the training step (ii), we fit a Gaussian Mixture Model (\textbf{GMM} in Section~\ref{subsec:gaussian_mixture}) to the set of latent codes of all the training data. In the sampling stage (iii), a random vector $z$ is first sampled from the GMM and then fed into the decoder network to produce a set of Gram matrices. Finally, we use the method~\cite{gatys2015texture} to synthesize a texture image from the generated Gram matrices. }\label{fig:main_pipeline}
\end{figure*}

Our method aims to explicitly generate texture features and then synthesize texture images from these features.
As discussed in Section~\ref{sec:introduction}, we choose the Gram matrix of the activation of convolutional feature maps as our texture feature, because it is shown to be a powerful representation and is widely used by many texture related tasks. That feature can be further used by a down-stream texture synthesis procedure~\cite{gatys2015texture} to generate a texture image.
The training and sampling pipelines are shown in Figure~\ref{fig:main_pipeline}.

\subsection{Framework Overview}
\label{subsec:gram_matrix_generation}

A Gram matrix computes the non-centered correlation between channels in a convolutional feature map. Conventionally, this feature map is extracted by the VGG-19 network pre-trained on object recognition \cite{simonyan2014very}. We use a set of Gram matrices $(G^{(1)},\dots,G^{(L)})$ computed from several layers in the network to specify the texture in our method.
We denote $M$ the mask of texture region in an image and compute the normalized gram matrix:
\begin{equation}
  \label{eq:gram_matrix}
  G^{(l)}_{ij} = \frac{1}{|M^{(l)}|} \sum_{k\in M^{(l)}} F^{(l)}_{k,i}F^{(l)}_{k,j},
\end{equation}
where $M^{(l)}$ is downsampled from $M$ accordingly and $F^{(l)}_{k,i}$ denotes the activation of the $i$-th channel at position $k$ of layer $l$ in the VGG-19 network. All texture features, $\{X_i = (G^{(1,i)}, \dots, G^{(L,i)})\}_{i=1}^n$, are extracted from the training images as our input data.

We base our generative model on the framework of Wasserstein Auto-Encoder (WAE)~\cite{tolstikhin2017wasserstein} as it is stable and produces better results.\footnote{We also experimented with another two generative frameworks: VAE and GAN. VAE tended to generate blurred images, while GAN was very unstable and easily collapsed to nonsense solutions.} The goal is to learn an auto-encoder which can reconstruct the input data and meanwhile constrain the distance between the empirical distribution of latent codes and the reference prior $p_Z$. This is achieved by introducing a reconstruction loss to the output and a divergence loss to the latent code (Figure~\ref{fig:main_pipeline} (i)).
In the sampling stage, a random noise vector is sampled from the prior distribution and then fed into the decoder network to produce a set of Gram matrices.
In the following subsection, we will elaborate the technical details of the three proposed components.

\subsection{Key Components}
\label{subsec:techincal_details}

\subsubsection{Gram Matrix Transformation}
\label{subsec:gram_matrix_transformation}

\begin{figure}
  \centering
  \includegraphics[width=0.98\linewidth]{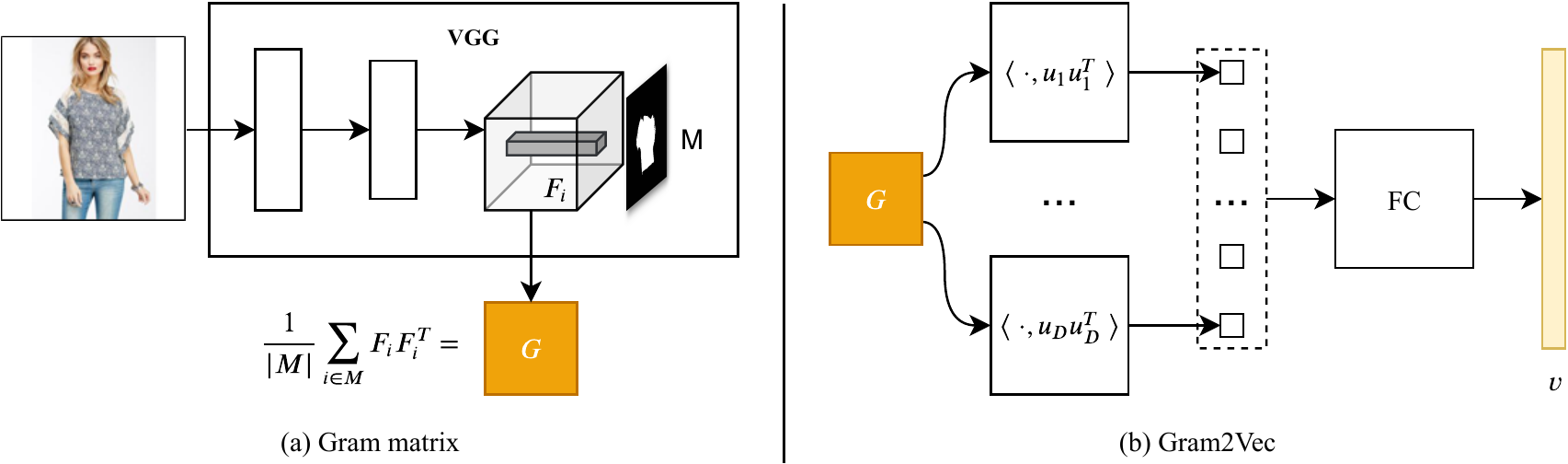}\\
  \caption{Gram matrix transformation. (a) shows the computation of normalized Gram matrix with mask. (b) shows the Gram matrix transformation (G2V). $\langle \cdot, \cdot\rangle$ denotes the inner product of two matrices. FC is a fully connected layer.}\label{fig:gram_transformation}
\end{figure}

Gram matrices can be very high-dimensional in deep layers of a convolutional network as their size is proportional to square of the number of channels. We introduce a transformation layer for the ease of reducing model size and computational complexity. A simple yet effective method is to use a fully connected layer to transform a Gram matrix into a vector. Recall that a Gram matrix is symmetric (Figure~\ref{fig:gram_transformation} (a)). Without loss of generality, the weight matrix in each filter of the FC layer is also symmetric and can be diagonalised. Thus, the FC transformation with a $d$ dimensional output can be reformulated as:
\begin{equation}
  \begin{aligned}
  & f_{FC}(G;W) = [\langle W^{(1)}, G\rangle, \dots, \langle W^{(d)}, G\rangle] \\
  & = \left[\sum_{j=1}^C\gamma^{(1)}_j \langle u^{(1)}_j {u^{(1)}_j}^T, G\rangle, \dots, \sum_{j=1}^C\gamma^{(d)}_j \langle u^{(d)}_j {u^{(d)}_j}^T, G\rangle\right]
  \end{aligned}
\end{equation}
or in general
\begin{eqnarray}
\label{eq:gram_transformation}
  f_{FC}(G;W(U,\Gamma)) = f_{FC}([\langle u_i u_i^T, G\rangle]_{i\in [D]}; \text{diag}(\Gamma)),
\end{eqnarray}
where $G\in \mathbb{R}^{C\times C}$ is a Gram matrix, $W = \{W^{(k)}\}_{k=1}^d \in \mathbb{R}^{d\times C\times C}$ is the weight of FC which can be parameterized by $d$ sets of eigenvectors $U = \{[u_1^{(k)}, \dots, u_C^{(k)}]\}_{k=1}^d \in \mathbb{R}^{d\times C\times C}$ and eigenvalues $\Gamma = \{[\gamma_1^{(k)}, \dots, \gamma_C^{(k)}]\}_{k=1}^d \in \mathbb{R}^{d\times C}$, and $D = C\cdot d$ is the number of combinations of subscript and superscript. Figure~\ref{fig:gram_transformation} (b) demonstrates the computation of Eq.~\ref{eq:gram_transformation}.

We assume that a small number of matrices $u_i u_i^T,~i\in [D]$ with $D \ll C\cdot d$ can represent most variance in the Gram matrix.\footnote{In the experiments, we set $D = 8\cdot C$, which reduces the number of parameters to 1/17  of the FC counterpart.} We denote the designed transformation layer by Gram2Vec (G2V). Its inverse transformation, denoted by Vec2Gram (V2G), is used in the generator to transform a hidden vector back into a Gram matrix.\footnote{The similar technique is used in bilinear pooling~\cite{gao2016compact} and second-order pooling~\cite{yu2018statistically}. However, their setups are designed for classification problems, which only requires a compression part. Besides, we do not append any normalization layer to the transformed vector and do not have any difficulty in optimization.} We employ the form of Eq.~\ref{eq:gram_transformation} and let the model learn the parameters in a data-driven manner.

\subsubsection{Recursive Auto-Encoder}
\label{subsec:recursive_autoencoder}

As the gram matrices are extracted from a sequence of feature maps from several layers of a convolutional neural network, they are closely correlated with each other. If we treat them as a union, there is a risk of generating uncorrelated Gram matrices although all of them seem to be realistic. This will cause difficulty in the following synthesis process and generate unrealistic textures.
To better model the correlations between Gram matrices, we propose to use a recursive auto-encoder (Figure~\ref{fig:main_pipeline}). In the encoder network for example, a stack of Recursive Units (RU) are used to encode the transformed Gram matrices layer-by-layer. In each RU, the transformed Gram matrix vector $v^{(l)}$ is first combined with the hidden state vector $h^{(l-1)}$ from the previous layer by $r$ fully connected (FC) layers. All the FC layers are preceded by ReLU~\cite{krizhevsky2012imagenet}.
Finally, the combined vector is added to $h^{(l-1)}$ to derive the hidden state vector $h^{(l)}$ for the current layer. The shortcut connection is used to ensure better gradient flow. An option in the direction of encoder remains to be determined: either bottom-up or top-down. After conducting more experiments, we find that the encoder in the bottom-up manner (as shown in Figure~\ref{fig:network} (a)) and the generator in the top-down manner produce samples with the best quality. We fix the directions of auto-encoder in all of our experiments.

\subsubsection{Gaussian Mixture Model in Latent Space}
\label{subsec:gaussian_mixture}

In the training stage, the latent code distribution is constrained by a divergence loss from the reference distribution $p_Z = N(0,I)$. In practice, such divergence is unlikely to be eliminated due to limited model capacity and optimization method. This may cause problems including: (1) unrealistic samples (which is unseen in the training data), (2) lack of diversity (e.g. mode collapse in GAN), in the sampling stage. In order to produce sampled results with better quality and diversity, we fit a Gaussian Mixture Model (GMM) to the latent codes of training data after the training of Gram-WAE-GAN (Figure~\ref{fig:main_pipeline} (ii)). Then in the sampling stage, we sample the random noise vector $z$ from the fitted GMM. The extra training time is almost negligible comparing to that of WAE. Kindly note that the original method, sampling from $N(0, I)$, can be considered as a special case of our method as there is only one component $N(0, I)$ in the GMM. Also note that the GMM can be applied to auto-encoder based generative methods and be incorporated into a standard network structure by adding a FC layer with weights of the square root of covariance matrices and biases of the means after $z\sim N(0,I)$.

\subsection{Training Objectives}
\label{subsec:training_objectives}
In our framework, we use the WAE-GAN variant~\cite{tolstikhin2017wasserstein} (as it generates better results according to the original paper), where a discriminator is learnt to distinguish between random vectors from the prior distribution and latent codes from the training data and squared L2 distance is used as the reconstruction loss. The whole autoencoder is called Gram-WAE-GAN.
We denote the mapping functions and parameters of the $Enc$, $Dec$ and $Dis$ networks by $f_{Enc}, \theta_{Enc}$, $f_{Dec}, \theta_{Dec}$ and $f_{Dis}, \theta_{Dis}$ respectively. The objective functions for the networks are summarized as follows:
\begin{equation}
  \begin{aligned}
     L_{rec} ={} & \mathbb{E}_{X\sim p_{data}} \|X - f_{Dec}(f_{Enc}(X))\|^2_2,\\
     L_{adv} ={} & \mathbb{E}_{Z\sim p_{Z}} \log(f_{Dis}(Z)) \\
     & + \mathbb{E}_{X\sim p_{data}} \log(1 - f_{Dis}(f_{Enc}(X))),\\
     L_{Enc} ={} & L_{rec} + \lambda\cdot L_{adv}, \\
     L_{Dec} ={} & L_{rec}, \\
     L_{Dis} ={} & - L_{adv}. \label{eq:losses}
  \end{aligned}
\end{equation}
A parameter $\lambda$ is used to control the tradeoff between reconstruction loss and adversarial loss. The networks are learned to minimize their corresponding losses (Eq.~\ref{eq:losses}).

\section{Experiments}
\label{sec:experimental_results}

\subsection{Experimental Setup}
\label{subsec:experimental_setup}

\noindent\textbf{Dataset.} We choose a subset of images from DeepFashion dataset \cite{liu2016deepfashion} with homogenous textures for the experiments in this paper. This dataset contains a large number of clothing images with comprehensive annotations. Besides the clothing we are interested in, each image also contains areas of human face, limbs and background. This is different from other texture datasets, such as the Oxford Describable Textures Dataset (DTD)~\cite{cimpoi14describing} and the Facades dataset~\cite{tylecek13}, in which the whole image is generally considered as a texture patch. These non-texture regions will contaminate the texture representation and thus should be excluded from the data.
To be specific, we first select the images with attributes containing keywords that represent a homogeneous texture such as ``dotted'', ``striped'', ``floral'', etc. Then we use the clothing masks extracted by \cite{zhu2017your} to further select texture regions out of whole images. Finally, this results in a subset of 13791 images. It is more challenging than other texture datasets in both aspects of quantity and diversity.

\begin{figure}
  \centering
  \includegraphics[width=0.98\linewidth]{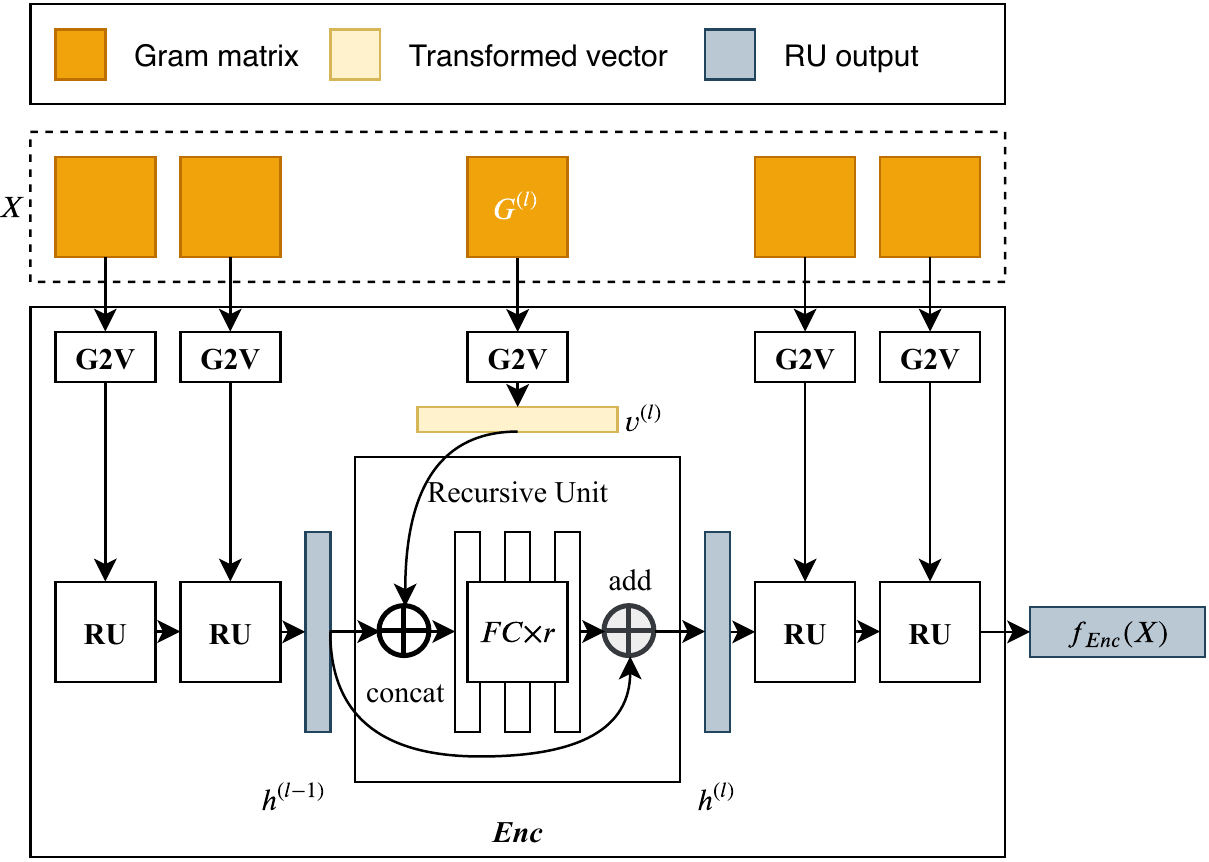}\\
  \caption{Recursive Auto-Encoder. This graph shows the structure of the encoder network. G2V denotes a Gram2Vec layer and RU denotes a recursive unit. The decoder (generator) network is a symmetric version of the encoder network. In a recursive unit. the hidden vector is first concatenated with the transformed vector computed from the current Gram matrix, and then passed through $r$ fully connected layers. All neurons are activated by ReLU before fed into FC, except the latent code (the last output).}\label{fig:network}
\end{figure}

\noindent\textbf{Network Architectures.} Our generator network is a recursive auto-encoder as described in Section~\ref{subsec:recursive_autoencoder}. The dimension of latent space is $d_e = 128$. In the recursive unit, the dimension of fc layer is $d_r = 512$ and the number of layers is $r = 2$. The architecture of encoder is illustrated in Figure~\ref{fig:network} and the generator is a symmetric version of the encoder. The discriminator is a 4-layer MLP with intermediate dimension $d_{dis} = 512$.

\noindent\textbf{Implementation Details.} In the training stage, a set of Gram matrices is extracted from the training samples and serves as the input data of our auto-encoder. The weight of adversarial loss is set to $\lambda_{adv} = 0.1$. We use ADAM \cite{kingma2014adam} optimization method with the setting of \cite{tolstikhin2017wasserstein} except for learning rate of 0.0001 and batch size of 64. After the training is finished, a Gaussian mixture model is fitted to all the latent codes of training samples. The number of components in the Gaussian mixture model is set to $n_c = 12$ which is determined by cross validation.
In the inference stage, the encoding network is ignored. A random vector is first sampled from the fitted Gaussian mixture model and fed into the generator (\ie decoder) to output a set of Gram matrices. These Gram matrices are then used by the texture synthesis algorithm \cite{gatys2015texture} to obtain a fashion image.

\noindent\textbf{Comparison Methods.} We choose three state-of-the-art methods, DistGAN \cite{tran2018dist}, PSGAN \cite{bergmann2017learning} and TextureGAN \cite{xian2018texturegan}, as the baseline methods. DistGAN is an improved GAN method which alleviates the gradient vanishing and mode collapse problems. Like vanilla GAN, this method learns to generate general images in the image space. PSGAN is specially designed for generating texture images. It outperformed previous methods on certain texture datasets. TextureGAN learns to synthesize texture images from sketch and texture patch given by user. Precisely, it is not a generative model like the previous two methods and ours. We choose it as a representative for user-guided texture synthesis. The qualitative and quantitative results are shown in the following subsection.

\subsection{Experimental Results}
\label{subsec:experimental_results}

\begin{figure*}
  \centering
  \includegraphics[width=0.98\linewidth]{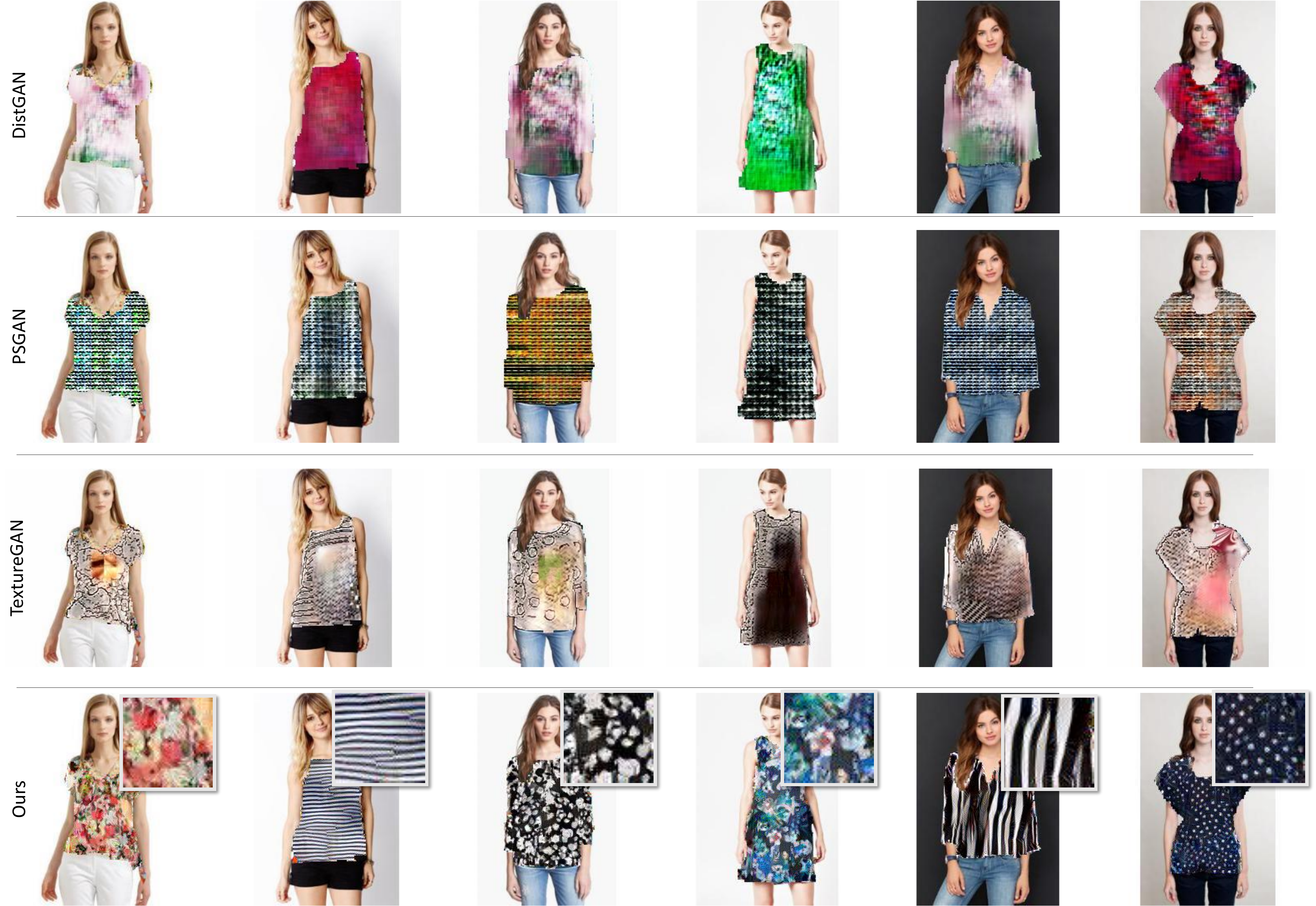}\\
  \caption{Qualitative results. (1) DistGAN~\cite{tran2018dist}, (2) PSGAN~\cite{bergmann2017learning}, (3) TextureGAN~\cite{xian2018texturegan} and (4) ours. The resolution of generated images is 256x256. Zoomed-in texture patches are shown in the top-right corner for our method.}\label{fig:baselines}
\end{figure*}

\paragraph{Qualitative Results} The generated samples of the baseline methods and ours are plotted in Figure~\ref{fig:baselines}. We can see that DistGAN tends to generate smooth color regions. Similar results can be observed in~\cite{zhu2017your} and~\cite{lassner2017generative} (Figure~\ref{fig:overview} (a-b)). These methods are designed for general images and fall short in generating texture details. PSGAN is able to generate high-frequency textures, but the quality of samples is not good enough. We used their public code and was able to reproduce their results on a smaller dataset (DTD). When applied to our dataset, the generated results cannot improve after enough iterations. Please note that we have tried multiplying the dimension of their model to ensure a fair comparison. We believe that this is due to the larger diversity in the dataset, as the original dataset used in PSGAN is relatively smaller than the one used in our experiments and contains only images sharing the same textural category and similar spatial structures. TextureGAN can extend the color of reference texture patches, but the results texture pattern is not preserved. Our method can generate textures with structures and varied colors like striped, dotted and floral patterns.

\begin{figure}
    \centering
      \includegraphics[width=0.98\linewidth]{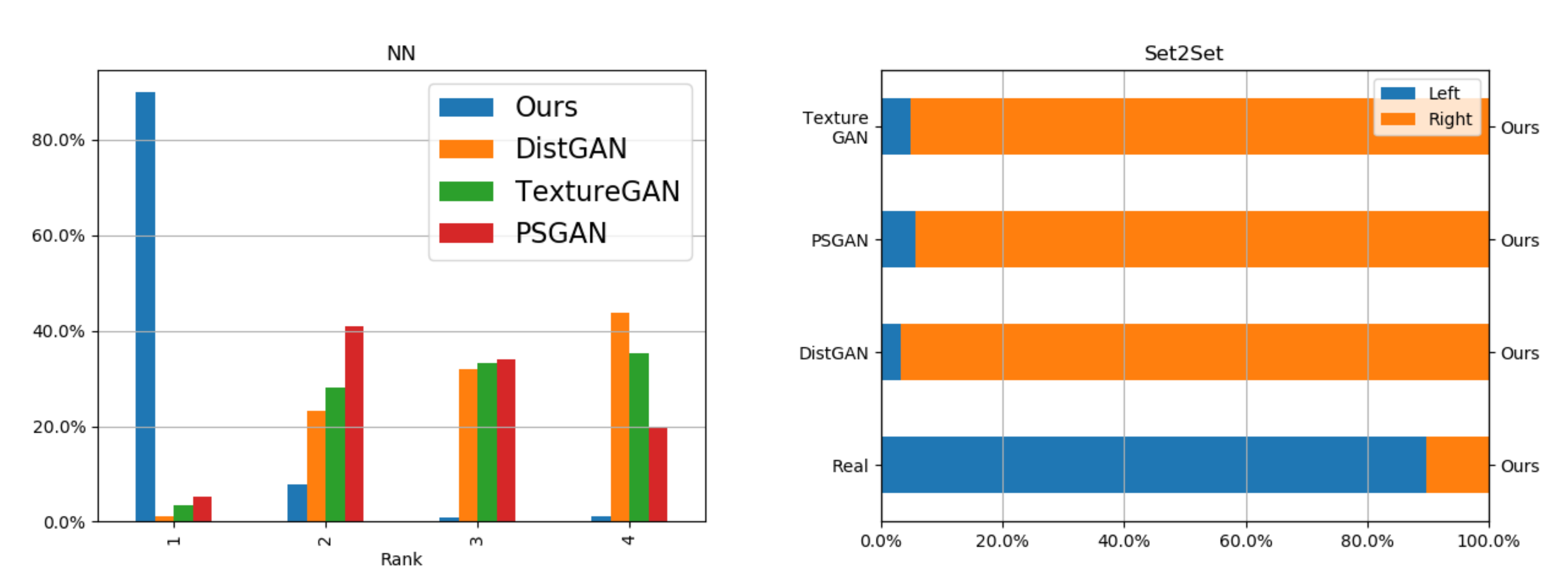}\\
      \caption{Results of user study: The first graph shows the ranking results for user study (NN). The second graph shows the comparison between our method and each baseline method (Set2Set).}\label{fig:user_study}
\end{figure}

\paragraph{User Study} We further conducted a user study to evaluate the quality of texture images generated from different methods. Unlike paired image-to-image synthesis (\eg super-resolution, and style transfer), the ground-truth output is unknown for each generated sample in our setting. It is also not clear how to compare the quality of generated textures from different categories. For example, one cannot compare a striped dress with a dotted dress directly. To make the comparison as fair and reasonable as possible, we designed two test settings for the user study: (1) \textit{Set2Set}: Each method randomly generates the same number of images to be grouped into a set. Each time we show two sets of images generated from two methods to users for comparison. The users are required to choose the one with better quality and richer diversity. (2) \textit{NN}: Each method randomly generates a large number of images to form a candidate set. Then we randomly select real images from test set as query images and find the nearest neighbor from each candidate set. Finally the nearest neighbors selected from all methods are shown to users for ranking. We perform the two tests respectively with 10 human raters and the statistical results are shown in Figure~\ref{fig:user_study}. Our method outperforms all baseline methods in terms of ranking among nearest neighbors and comparison between sets.

\paragraph{FID Score} Besides human ratings, we also compute the Fr\'echet Inception Distance (FID) \cite{heusel2017gans} between generated data and real data for the three generative methods: DistGAN, PSGAN and ours. In the literature of GAN, FID is widely used to evaluate the realism and diversity of the generated images. Lower FID means closer distance between the generated data and the real data. The numeric results are listed in Table~\ref{tab:fid_scores}. Our method outperforms the other methods by a large margin, which aligns with the visual results.

\begin{table}[h]
  \centering
  \begin{tabular}{llc}
      \hline
      & \textbf{Method} & \textbf{FID} \\
      \hline
      \multirow{3}{*}{Baseline} & DistGAN~\cite{tran2018dist} & 41.97 \\
      & PSGAN~\cite{bergmann2017learning} & 77.10 \\
      & TextureGAN~\cite{xian2018texturegan} & 44.38 \\
      \hline
      \multirow{3}{*}{\makecell{Ablation\\Study}} & FC transformation & \textit{\textbf{37.32}} \\
      & MLP structure & 45.72 \\
      & No GMM sampling & 40.83 \\
      \hline
      & Ours & \textbf{37.74} \\
      \hline
  \end{tabular}
  \caption{FID scores (the lower the better). Our method outperforms the other baseline methods by a large margin, which aligns with the visual results. For the variants introduced in Section~\ref{subsec:ablation_study}, our method achieves comparable results with the FC transformation method while using significantly less parameters. Our method also outperforms the other two methods which shows the effectiveness of the proposed components in generating realistic and diversified results.}\label{tab:fid_scores}
\end{table}

\subsection{Ablation Study}
\label{subsec:ablation_study}

In this section, we conduct more experiments to investigate the effectiveness of each proposed component as ablation study.

\begin{figure}
  \centering
  \includegraphics[width=0.98\linewidth]{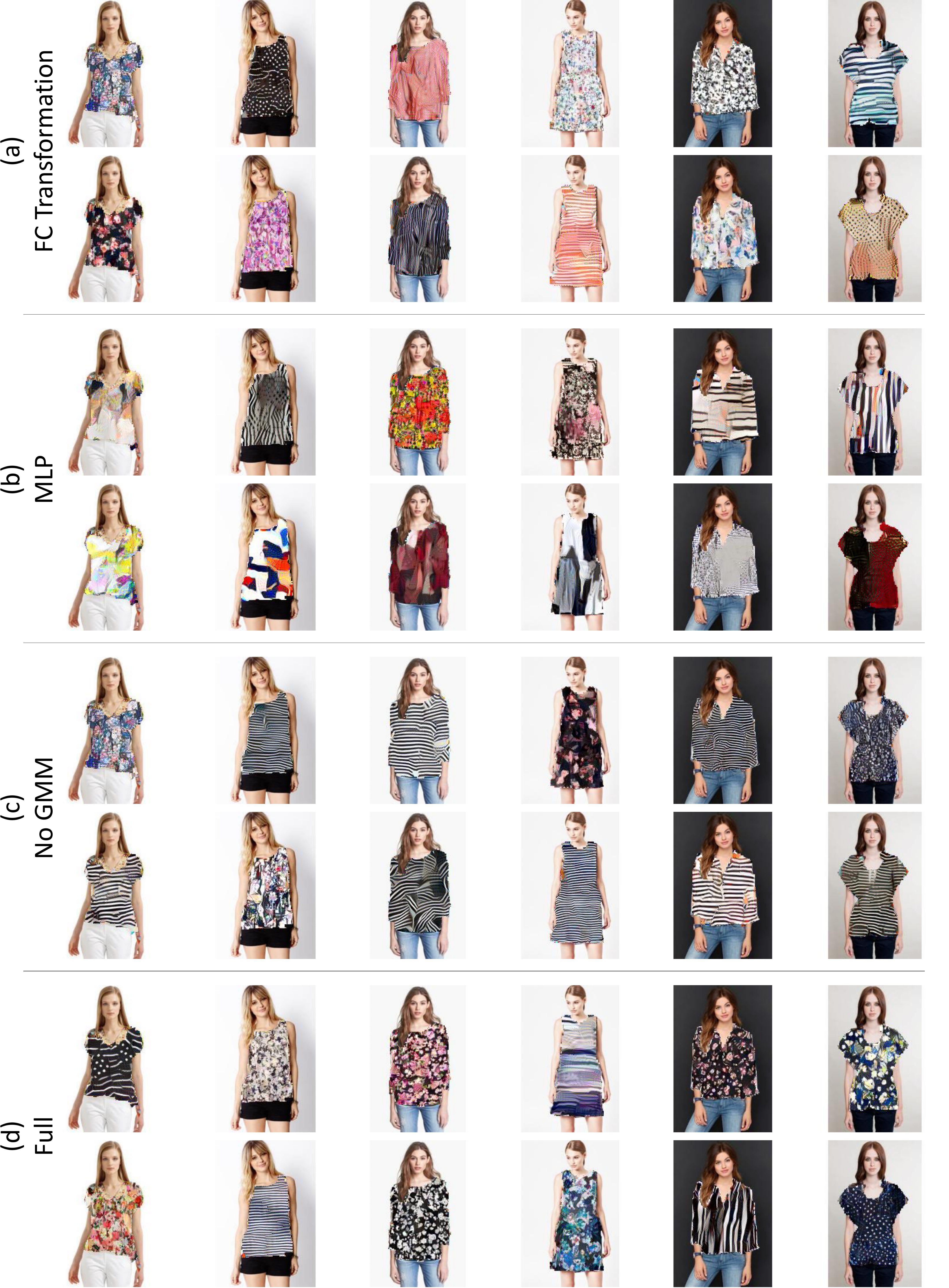}\\
  \caption{Generated results for ablation study. From top to bottom: (a) our method using fully connected layer as the Gram matrix transformation, (b) our method using MLP auto-encoder, (c) our method using noises from $N(0,I)$, (d) our full method.}\label{fig:ablation_study}
\end{figure}


\paragraph{Gram Matrix Transformation} We replace each G2V (and V2G) layer by a fully connected layer to transform a Gram matrix into (and from) a vector and rerun the same training procedure. The FID of the FC variant is 37.32, which is slightly better than the proposed method (37.74). The generated samples are listed in Figure~\ref{fig:ablation_study} (the first and the fourth groups). We can see that our method using Gram matrix transformation can generate results with comparative quality, while it uses significantly less parameters than the FC variant.

\begin{table}[h]
  \centering
  \begin{tabular}{lcc}
    \hline
    \textbf{Method} & Gram matrix & FC \\\hline
    \textbf{\# params} & 10.8M & 184M \\
    \hline
  \end{tabular}
  \caption{The number of parameters of our methods using Gram matrix transformation and fully connected layer.}\label{tab:ablation_gram_matrix_transformation}
\end{table}


\paragraph{Recursive Network} We conduct experiments using MLP structure in the auto-encoder to evaluate the effectiveness of the designed recursive structure. Specifically, both the encoder and generator consists of a 4-layer MLP, and the Gram matrix transformation layers are retained between the input (output) and the main body of network. The number of parameters is adjusted to be comparable with that of our method using recursive units. From Figure~\ref{fig:ablation_study} (the second group) we can see that some samples contain different color blocks or inconsistent textural patterns. The inconsistence in the texture feature indicates that the correlation between Gram matrices from different layers is not retained. We also observe constantly higher optimization error in the synthesis process. The synthesizer has difficulty to draw a texture image that complies with all the Gram matrices at the same time. The FID of this variant raises significantly to 45.72. Our recursive structure shows its advantage to modeling the correlation between Gram matrices.

\begin{figure}
  \centering
  \includegraphics[width=0.98\linewidth]{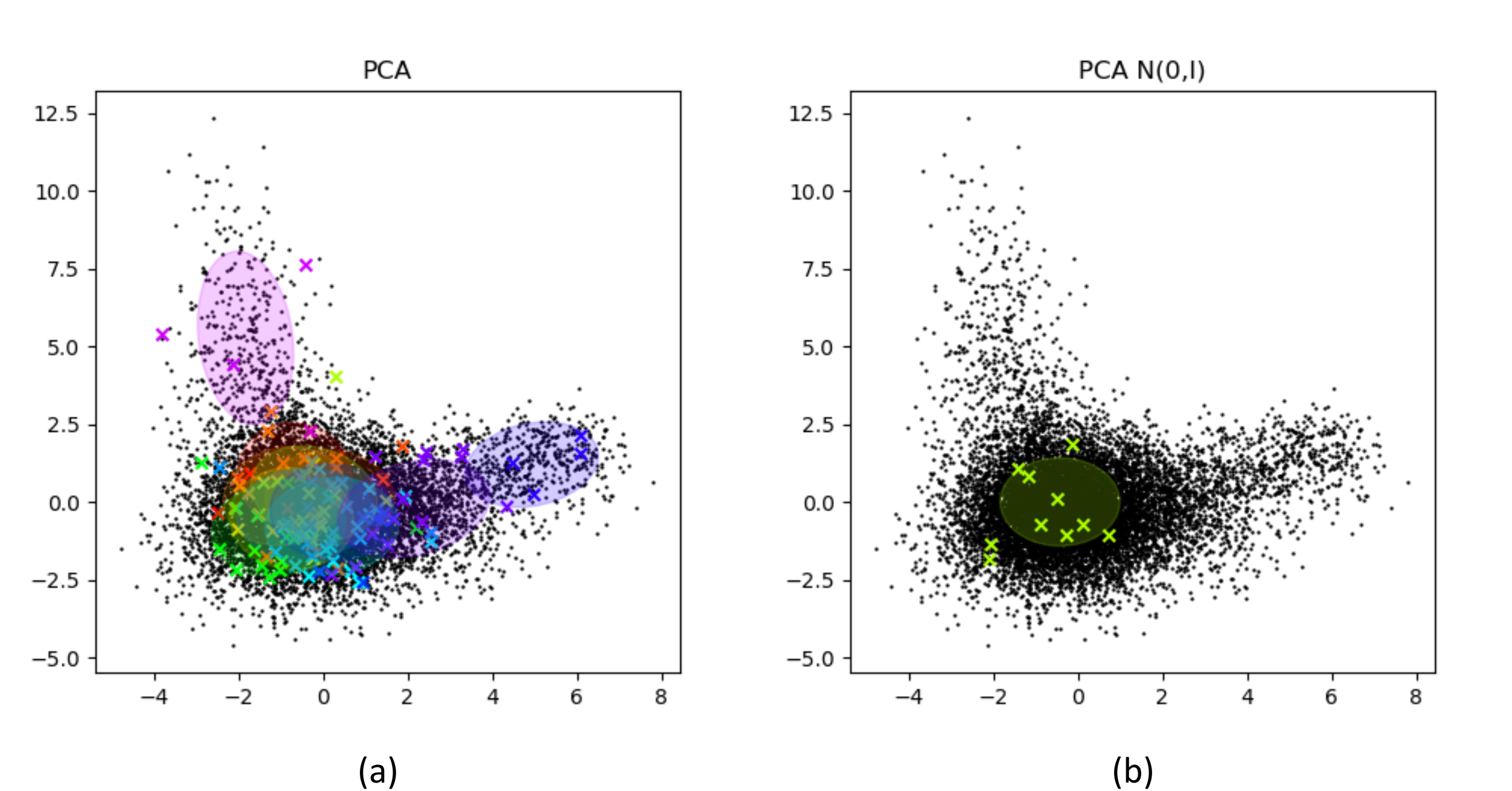}\\
  \caption{Embedding of latent codes. The black dots are the embedding of real latent codes. The colored crossings are the embedding of sampled noises. Each color corresponds to a component in GMM. The ellipses illustrate the projected covariance matrices. (a) shows the embedding of samples from the fitted GMM. (b) shows the embedding of samples from $N(0,I)$.}\label{fig:ablation_gmm}
\end{figure}

\paragraph{Gaussian Mixture Model} In this experiment, we use the same recursive auto-encoder after training. When doing inference, instead of sampling from the Gaussian mixture model, we sample noises from the standard Gaussian distribution $N(0, I)$. This is the conventional approach used by many generative models. From Figure~\ref{fig:ablation_study} (c), we can see that this variant tends to generate similar striped texture pattern. However, our full method can generate textures with varied structures and colors. We also plot the 2d PCA embedding of the latent codes of training samples and random noises sampled from GMM and $N(0, I)$ in Figure~\ref{fig:ablation_gmm}. The noises sampled from $N(0, I)$ only cover a small portion of real samples, while the noises sampled from GMM can better fit to the real distribution. The FID score of this variant is 40.83 which indicates that our proposed sampling method can generate images with more diversity.

\section{Conclusion}
\label{sec:conclusion}
In this work, we have proposed a novel generative model for synthesizing fashion textures from scratch. Our approach explicitly generates a set of Gram matrices as the texture feature and then synthesizes a texture image from that feature. Three key components are designed for the challenges of generating texture images. We evaluated our method on a large and diverse texture dataset. The experimental results validate that our method is capable of generating realistic and diverse texture images. Furthermore, it outperforms previous methods with respect to visual quality, distance metric (FID) and user preference by a large margin. 

{\small
\bibliographystyle{ieee_fullname}
\bibliography{egbib}
}

\end{document}